# Machine Learning Technique Based Fake News Detection

Biplob Kumar Sutradhar[1, a)], Md. Zonaid[1, b)], Nushrat Jahan Ria[1, c)] and Sheak Rashed Haider Noori[1, d)]

*Author Affiliations*
[1] *Daffodil International University, Dhaka, Bangladesh*

*Author Emails*
[a)] biplob15-3923@diu.edu.bd
[b)] zonaid15-3927@diu.edu.bd
[c)] nushratria.cse@diu.edu.bd
[d)] drnoori@daffodilvarsity.edu.bd

**Abstract:** False news has received attention from both the general public and the scholarly world. Such false information has the ability to affect public perception, giving nefarious groups the chance to influence the results of public events like elections. Anyone can share fake news or facts about anyone or anything for their personal gain or to cause someone trouble. Also, information varies depending on the part of the world it is shared on. Thus, in this paper, we have trained a model to classify fake and true news by utilizing the 1876 news data from our collected dataset. We have preprocessed the data to get clean and filtered texts by following the Natural Language Processing approaches. Our research conducts 3 popular Machine Learning (Stochastic gradient descent, Naïve Bayes, Logistic Regression,) and 2 Deep Learning (Long-Short Term Memory, ASGD Weight-Dropped LSTM, or AWD-LSTM) algorithms. After we have found our best Naive Bayes classifier with 56% accuracy and an F1-macro score of an average of 32%.

## INTRODUCTION

The consumption of news among people is increasing because of low cost and easily accessible technology. Now it's becoming easier to share news instantly via social media as people spend most of their time on social media. Thus, fake news also takes this advantage of technology and can have large scale negative effects on political (USA election, 2015), social and economic levels [1]. Sharing news without checking the quality of content is also making news quality questionable and news losing its ground universally. Spammers use social media to share their clickbait and influence traffic to fake news [2]. Social media platforms are very powerful in creating spreading satire or absurdity, biased opinions and manipulating mindsets [3]. Some traditional computational techniques also can be found and used to target specific text as hoax on basis of the textual content. However, the majority of these just check these websites are "PolitiFact" and "Snopes". Humans maintain these website repositories and these consist of a list of websites classified as ambiguous and fake [4].

Nevertheless, this list does not classify all the categories of news; it is specifically for political news. Humans' ability to detect fake news content is 54% [5]. Every day the number of news content all over the world is increasing rapidly, and it's becoming impossible for humans to classify all the news that get published. On the other hand, the process is also very time consuming.

To detect fake news from all categories NLP can play a major role. So, in this research we used NLP with feature classification to detect fake news. ML and DL algorithms both are tested by us. In the later section, we presented related work and figuring out the previous works' gap, and we have presented our own approaches to increase outcome of fake news detection.

# LITERATURE REVIEW

Fake news detection is getting more and more attention nowadays by scholars from different backgrounds. To solve this problem authors used Single Modal and Multi-modal detection approaches.

In fake news detection tasks, the main challenge for researchers is how to distinguish news according to features, how data should be classified, and which classification will give better output. A work based on A linguistic model to search language driven features for classification done before. For a news article, this model can find grammatical, syntax-based features, word density, and sentimental features. Then for classifying news Choudhary et al. [6] used Sequential learning models in this research. Sometimes only linguistic features and classification don't go well for different categories of news. Different categories of news have a specific linguistic pattern. Thus, L. Wu et al. [7] took another approach, user behavior analysis, propagation of the message. Shen et al. [8] used social engagements on social media, syntactic and Semantic features were taken into consideration for better results. Another group has taken into consideration the readability matrix. Some authors presented that previous works are very specific event dependent. They had tried to come up with a solution that could be even adversarial (event independent). They used visual features also to classify news. To extract textual features, they used modified text-CNN with various size filter windows, and for visual features they also used CNN. Their approach was very effective, when the news content had images and videos related to this news. Authors [10] showed that the quality of feature extractions from a Mask-RCNN model can be improved by using multi-task learning. Sami et al. [9] proposed a new loss function that is a variant of the focal loss.

The machine learning ensemble approach works on specific article domain's that have unique textual structures. L. Wu et al. [7] encountered features like the LIWC features. For classification they used SVM, LR, MP and KNN models. The research was conducted in ISOT Fake News Dataset. 98% accuracy was achieved by boosting classifiers algorithm and multilayer perceptron, Perez-LSVM. Linear SVM, bagging classifiers. Iftikhar Ahmad et al. [11] proposed ML ensemble approach to classify news articles automatically. Their research looked at numerous linguistic qualities that may be utilized to tell the difference between phony and authentic content. They trained a mixture of distinct ML algorithms using diverse ensemble approaches and evaluated their performance on four real-world datasets using those attributes. In compared to individual learners, their suggested ensemble learner technique outperformed individual learners in an experiment.

Naive Bayes Classifier (NBC) is also used for spam filtering. It is based on Bayes' theorem, which assumes substantial (native) independence between the characteristics. Mykhailo Granik et al. [19] demonstrated a simple method for detecting bogus news using a NBC. NBC was turned into a software system and it tested a collection of Facebook news posts. Around 74% accuracy was gained by this classifier on the test set. The result was pretty much satisfying at that time. In the article the authors discussed about how their finding may help to improve the performance of the classifier. They suggested Artificial Intelligence approaches to detect false news through their findings. S. Gilda et al. [12] suggested TF-IDF model which is quite good at classifying articles from untrustworthy sources. Syntactical structure frequency (probabilistic context free grammars, or PCFGs), a feature union and bi-gram frequency are used in the TF-IDF model. To detect 'fake news,' or news items that are deceptive and come from untrustworthy sources they studied natural language processing techniques to detect. A Signal Media dataset as well as a list of sources from OpenSources.co was used in their work. On a dataset of roughly 11,000 articles, term frequency-inverse document frequency (TF-IDF) bi-gram detection and probabilistic context free grammar (PCFG) detection was used. To classify their dataset, they used SGD GB, SVM, BDT, and RF. They discovered that feeding TF-IDF of bi-grams into an SGD model correctly detects non-credible sources 77.2 percent of the time, with PCFGs having little influence on recall.

According to [28], variational auto-encoders rely on variational Bayesian inference for density approximation. Jeremy Howard et al. [17] stated that AWD-LSTM is a state-of-the-art language model, with various tuned dropout hyperparameters with a regular LSTM. Zein Shaheen et al. [18] also used this as an encoder, with extra layers added on top for the classification task. In their works, this model is the best fit for text classification.

# DATASET AND DATA PREPROCESSING

The fake news dataset used in this research consists of a training set, which is a combination of "english_dev.csv" with 364 rows and "english_training.csv" with 900 rows, totaling 1264 rows and from our own collection. The test set comprises 612 rows. Both the training and test datasets contain four attributes: "public_id", "title", "text", and "our rating". The "text" column contains the full news article, the "title" is the headline, and the "our rating" column indicates the veracity of the news (true, false, partially false, other)[16]. Note that the test dataset does not include a "our rating" column. Here a snippet of our train dataset given below

[Dataset snippet table showing public_id, text, title, our rating columns with 1264 rows × 4 columns]

Fig. 1. Shows a snippet of the dataset used in this research

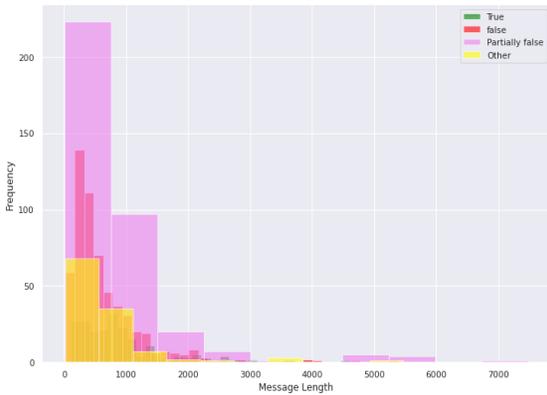

Fig. 2. Dataset word length plot

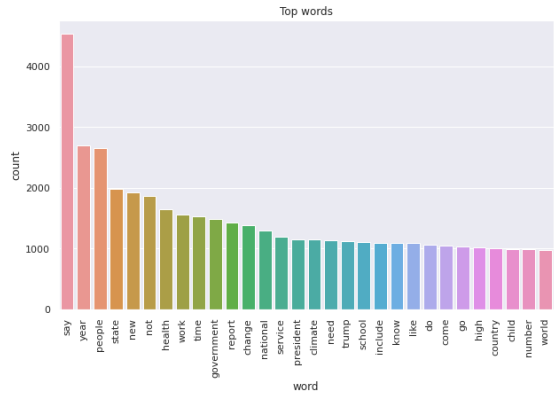

Fig. 3. Dataset top words bar plot

### A. CLASSES DEFINITIONS

Our rating column has 4 classes. A brief description and count class figure of these classes is given below.

- **False** - That means news is completely false and verified by fact check services.
- **True** - This means that news is true. News follows a true pattern.
- **Partially false** - The news may contain a mixture of true and false combination events. Also, information manipulated, misdescribed etc.
- **Other** - This is not news content or just a public user post or just describing a product review.

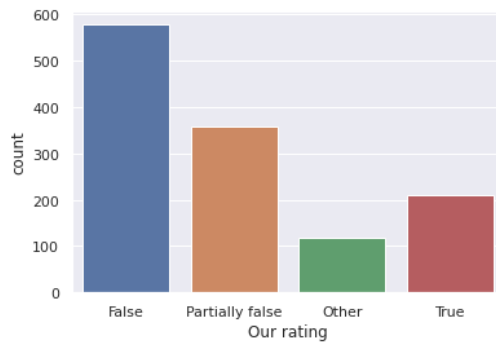

Fig. 4. Count class

## B. DATA PREPROCESSING

Cleaning and filtering data allows the model to fit. e.g., remove any URL, email, tag, non ascii, to lowercase, punctuation, stop words, lemmatize, replace nan value to blank string. We have removed constant e.g., number, punctuation, word length one and two characters. To achieve this, we have used python regular expressions, python external library text-hammer and natural language toolkit (nltk) tools.

Changing our ratting column value False is 0, True is 1, Partially false is 2 and Other is 3. Then concatenate the title and text column called fullmsg. Given below 5 is the after-preprocessing step.

| index | our rating | |
|---|---|---|
| 0 | 0 | fin,one thousand, five hundred,passenger,use,mobile,phone,start,next,week,distract,drive,cause,... |
| 1 | 2 | missouri,lawmakers,condemn,las,vegas,shoot,missouri,politicians,make,statements,mass,shoot,l... |
| 2 | 2 | cbc,cut,donald,trump,home,alone,somenumber,cameo,broadcast,home,alone,somenumber,lose,... |
| 3 | 0 | obama,daughters,catch,camera,burn,us,flag,dc,protest,iolgue,things,take,turn,worse,riot,police,fir... |
| 4 | 0 | leak,visitor,log,reveal,schiff,somenumber,visit,epstein,isle,secret,epstein,schiff,share,long,history,... |
| 5 | 3 | k,b,governor,secretary,state,georgia,take,money,china,steal,election,trump,nation,update,eight hu... |
| 6 | 0 | fda,shock,study,cells,use,vaccines,contaminate,serious,viruses,include,cancer,november,somenu... |
| 7 | 0 | israel,hit,beirut,nuclear,missile,trump,lebanese,govt,confirm,continually,update,veterans,today,trum... four,kilotons,tnt,equivalent,use,linear,extrapolation,tianjin,involve,somenumber,tonnes,ammonium... |
| 8 | 0 | obama,daughters,catch,camera,burn,us,flag,dc,protest,show,antiamerican,sentiment,surprise,pre... |
| 9 | 0 | field,human,cage,discover,caruthers,california,video,fema,camp,portable,human,cage,real,screen... |

Show 10 per page    1  2  10  100  120  127

Fig. 5. After preprocessing title and text

## C. DATA PREPROCESSING

We cannot directly use raw data of our models. It is not efficient data for our machine learning modeling. We have to transform that raw data into some features extracted from. In our work we have used TF-IDF, CountVectorizer, Word2vec and Fast.ai default tokenization(tok_tfm) [14][27].

1. **"TF-IDF"**: For TF-IDF, we have used this tool from the scikit-learn v1.0.2 Python library feature_extraction.text module.
2. **"CountVectorizer"**: This vectorizer turns a set of text documents into a document count metric. To transform text to counter vectors, we have used the scikit-learn v1.0.2 library in Python.
3. **"Word2vec"**: For converting words into real numbers, we have used word2vec from the Gensim library. Using Word2vec we generate word embeddings.
4. **"Fast.ai default tokenization"**: The Fast.ai automatically does some additional steps like tokenization and numericalization. So that Fast.ai AWD_LSTM will work with that. After preprocessing text (text is the title + text column) concatenate new column name text. AWD-LSTM is the base of FastAI's text models [15]. FastAI can auto tokenized text columns and make them ready to work with models [14].

## METHODOLOGY

To effectively detect fake information online, researchers must apply appropriate data processing techniques when working with below models.

In this research, we have used some NLP and deep learning techniques to conduct our research (Naive Bayes Classifier, Logistic regression, Stochastic gradient descent, Word2vec using gensim library LSTM, AWD_LSTM). Before applying these techniques, we have preprocessed our data with python external library which are already mentioned in data preprocessing section. We have mentioned step by step by process

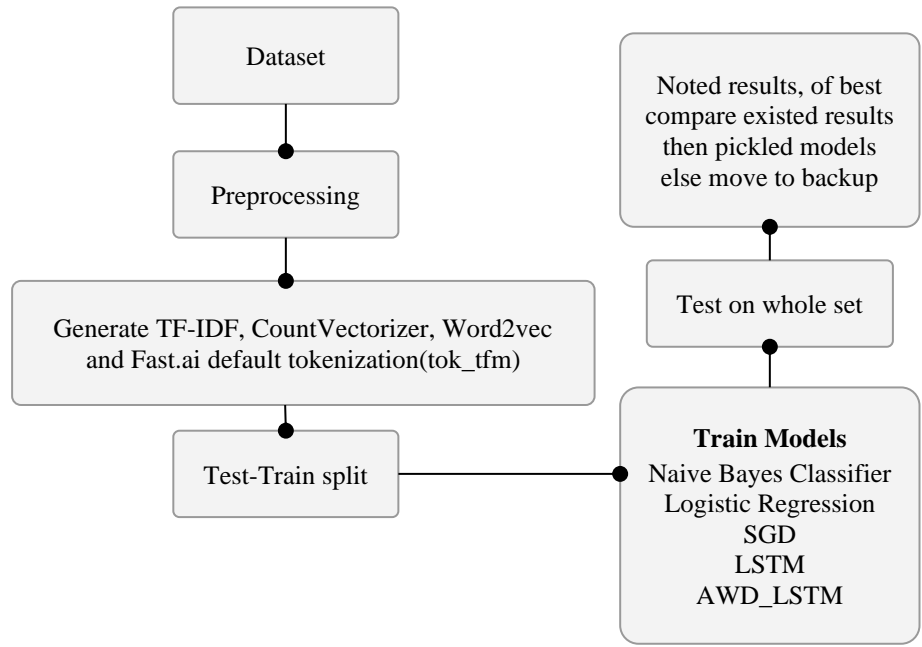

Fig. 6. Research test pipeline representation

## RESEARCH MODELS

In this research, we implemented below models to evaluate the veracity of news articles. In this below there are research models.

*1. Machine learning*

1. **"Naive Bayes (Using Count vectorizer Features)"**: The bias theorem is the foundation of the NBC. We have used scikit-learn to get naive bayes classifiers.
2. **"Logistic regression (Using word level tf-idf Features)"**: This is a supervised learning model. This means something where we use labeled data. It is also a classification model. It uses a sigmoid function. We got this model from scikit-learn.
3. **"Stochastic gradient descent (Using word level tf-idf Features)"**: Stochastic gradient descent (SGD) is an optimization algorithm that works iteratively. It is well suited for neural network optimization. We used this algorithm from the scikit-learn python library.

*2. Deep learning*

1. **"LSTM (Word2vec using gensim library)"**: Gensim library is a NLP library for python. We have used word2vec.And it is a more robust embedding method. That means we have used a pre-trend vector. We have used 'glove-wiki-gigaword-100' 100 dimensions of the vectors.
2. **"AWD_LSTM (Fast.ai)"**: This model used in this research is inspired from "Regularizing and Optimizing LSTM Language Models" [15]. It employs a drop LSTM that uses drop connections on 'hidden-to-hidden' weights as a method of returning to the usage table. Drop connect is a generalization of different types of drop out for regularization of large fully connected layers within a neural network.

## EXPERIMENT RESULT

We conducted three ML and two DL experiments and found that the Naive Bayes classifier (NBC) was the best performing machine learning algorithm, with a test data F1-macro average of 32% and accuracy of 56%.

TABLE 1. contains all algorithms accuracy and F1-macro average comparisons and TABLE 2, 3, 4, 5, 6 are detailed results of each algorithm. And last section CONFUSION MATRIX of each algorithm.

**TABLE 1.** Accuracy and f1-macro avg comparison report

| Algorithms | Accuracy | F1-macro avg |
|---|---|---|
| Naive Bayes Classifier | 56% | 32% |
| Logistic regression (LR) | 52% | 26% |
| SGD | 51% | 27% |
| LSTM | 39% | 23% |
| AWD_LSTM | 49% | 29% |

Naive bayes incorporate strong independence assumptions in datasets so in this task, this algorithm works well among other 4 algorithms (Logistic regression (LR), SGD, LSTM, AWD_LSTM).

## A. NAIVE BAYES CLASSIFIER (USING COUNTVECTORIZER FEATURES)

**TABLE 2.** Results of naive bayes

| Class name | Precision | Recall | F1-macro avg |
|---|---|---|---|
| False | 59% | 91% | 72% |
| True | 66% | 17% | 27% |
| Partially false | 26% | 34% | 30% |
| Other | 0% | 0% | 0% |

In our experiment for NBC, we have used the MultinomialNB() function from sklearn.naive_bayes library v1.0.2 to predict our classes.

## B. LOGISTIC REGRESSION (USING WORD LEVEL TF-IDF FEATURES)

**TABLE 3.** Results of logistic regression

| Class name | Precision | Recall | F1-macro avg |
|---|---|---|---|
| False | 59% | 90% | 71% |
| True | 64% | 11% | 19% |
| Partially false | 13% | 21% | 16% |
| Other | 0% | 0% | 0% |

Logistic regression [20] is used for the binary classification. However, we have tried for our experiment. In our experiment, we have used linear_model from sklearn library for Logistic Regression model. Our settings parameter for model is given here. Parameter C is 100.0 and solver is liblinear and other parameter values are default to sklearn library v1.0.2.

## C. STOCHASTIC GRADIENT DESCENT (USING WORD LEVEL TF-IDF FEATURES)

**TABLE 4.** Results of stochastic gradient descent

| Class name | Precision | Recall | F1-macro avg |
|---|---|---|---|
| False | 59% | 86% | 70% |
| True | 62% | 14% | 23% |
| Partially false | 13% | 23% | 17% |
| Other | 0% | 0% | 0% |

In our research, we have used SGDClassifier [22] from sklearn.linear_model library v1.0.2. Also the parameter for random_state is 42.

## D. LSTM (WORD2VEC USING GENSIM LIBRARY)

TABLE 5. Results of LSTM

| Class name | Precision | Recall | F1-macro avg |
|---|---|---|---|
| False | 59% | 86% | 70% |
| True | 62% | 14% | 23% |
| Partially false | 13% | 23% | 17% |
| Other | 0% | 0% | 0% |

LSTM networks have been successfully used for various tasks including text classification [24], question answering [25], and machine translation [26]. In our research we have used the Sequential function from tensorflow.keras.model's library v2.8.2 in our experiment. The fit epochs parameter for this algorithm is settled 25.

## E. AWD_LSTM (FAST.AI)

TABLE 6. Results of AWD_LSTM

| Class name | Precision | Recall | F1-macro avg |
|---|---|---|---|
| False | 75% | 17% | 28% |
| True | 58% | 80% | 68% |
| Partially false | 10% | 18% | 12% |
| Other | 8% | 6% | 7% |

We have used AWD_LSTM from fastAI library v2.6.3 in this experiment. In the text_classifier_learner function our setting parameter for drop_mult is 0.5 and epoch for fit is 10.

## CONFUSION MATRIX

In the confusion matrix we have found that the first three machine learning models (naive bayes, Logistic regression, SGD) predict more accurately when the data is already false. As our dataset is imbalanced, the False labeled data is more than other classes thus, during the training season, the model learned better to predict False data.

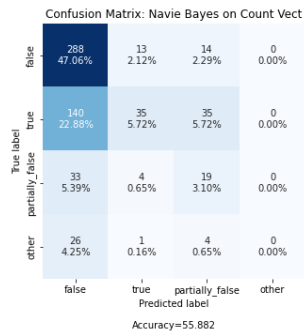

Fig. 7. NBC when using Countvectorizer features

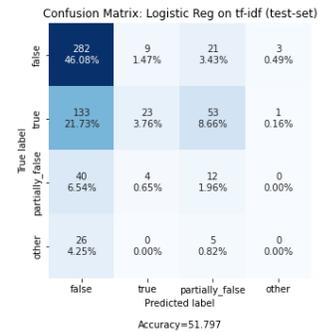

Fig. 8. LR model when using word level tf-idf features

Fig. 9. SGD (Using word level tf-idf Features)

Fig. 10. LSTM (Word2vec using gensim library)

Fig. 11. AWD_LSTM (Fast.ai)

However, our deep learning model doesn't do well for this fake news detection problem as deep learning models need more data to predict more accurately. Hence, the results when deep learning models are used, didn't perform well.

## CONCLUSIONS

One of the most dangerous risks on social media recently has emerged is false news. Malicious actors may use fake news to influence people's choices and judgments on crucial everyday activities including financial markets, healthcare, online shopping, education, and even presidential elections. It is a very important yet difficult undertaking for both industry and academics to automatically recognize online fake news [23]. We tried out five ML and DL models to classify fake news detection. After all, we have found Naive Bayes Classifier was found to be the most effective model for fake news classification on our research, achieving a F1-macro average of 32% on the latest test results. NBC is the best model for this fake news classification with F1-macro average 32% on the last release of test result. In future, we will work on more new research models to be able to know which model is best for this fake news classification problem.

## ACKNOWLEDGMENT

We would like to thank Daffodil International University's NLP and ML Research Lab for their invaluable support and assistance during the course of this research project. We are deeply appreciative of the university's collaboration and support, which played a crucial role in the success of this research.

## REFERENCES

1. W Kogan, S., Moskowitz, T. J., & Niessner, M. (2019). Fake news: Evidence from financial markets. Available at SSRN, 3237763.
2. Wong, J. (2016). Almost all the traffic to fake news sites is from facebook, new data show. The Medium.

3. Lazer, D. M., Baum, M. A., Benkler, Y., Berinsky, A. J., Greenhill, K. M., Menczer, F., ... & Zittrain, J. L. (2018). The science of fake news. Science, 359(6380), 1094-1096.
4. Asr, F. T., & Taboada, M. (2019). Misinfotext: a collection of news articles, with false and true labels.
5. Krishnamurthy, G., Majumder, N., Poria, S., & Cambria, E. (2018). A deep learning approach for multimodal deception detection. arXiv preprint arXiv:1803.00344.
6. Choudhary, A., & Arora, A. (2021). Linguistic feature-based learning model for fake news detection and classification. Expert Systems with Applications, 169, 114171.
7. Wu, L., & Liu, H. (2018, February). Tracing fake-news footprints: Characterizing social media messages by how they propagate. In Proceedings of the eleventh ACM international conference on Web Search and Data Mining (pp. 637-645).
8. Shen, H., Ma, F., Zhang, X., Zong, L., Liu, X., & Liang, W. (2017). Discovering social spammers from multiple views. Neurocomputing, 225, 49-57.
9. Sami, M. T., Yan, D., Joy, B. R., Khalil, J., Cevallos, R., Hossain, M. E., ... & Zhou, Y. (2022, November). Center-Based iPSC Colony Counting with Multi-Task Learning. In *2022 IEEE International Conference on Data Mining (ICDM)* (pp. 1173-1178). IEEE..
10. Sami, M. T., Yan, D., Huang, H., Liang, X., Guo, G., & Jiang, Z. (2021, December). Drone-Based Tower Survey by Multi-Task Learning. In *2021 IEEE international conference on big data (big data)* (pp. 6011-6013). IEEE.
11. Ahmad, I., Yousaf, M., Yousaf, S., & Ahmad, M. O. (2020). Fake news detection using machine learning ensemble methods. Complexity, 2020.
12. Gilda, S. (2017, December). Notice of Violation of IEEE Publication Principles: Evaluating machine learning algorithms for fake news detection. In 2017 IEEE 15th student conference on research and development (SCOReD) (pp. 110-115). IEEE.
13. Islam, M. R., Liu, S., Wang, X., & Xu, G. (2020). Deep learning for misinformation detection on online social networks: a survey and new perspectives. Social Network Analysis and Mining, 10(1), 1-20.
14. Howard, J., & Gugger, S. (2020). Fastai: a layered API for deep learning. Information, 11(2), 108.
15. Merity, S., Keskar, N. S., & Socher, R. (2017). Regularizing and optimizing LSTM language models. arXiv preprint arXiv:1708.02182.
16. Shahi, G. K., Dirkson, A., & Majchrzak, T. A. (2021). An exploratory study of covid-19 misinformation on twitter. Online social networks and media, 22, 100104.
17. Shahi, G. K., & Nandini, D. (2020). FakeCovid--A multilingual cross-domain fact check news dataset for COVID-19. arXiv preprint arXiv:2006.11343.
18. Howard, J., & Ruder, S. (2018). Universal language model fine-tuning for text classification. arXiv preprint arXiv:1801.06146.
19. Shaheen, Z., Wohlgenannt, G., & Filtz, E. (2020). Large scale legal text classification using transformer models. arXiv preprint arXiv:2010.12871.
20. Granik, M., & Mesyura, V. (2017, May). Fake news detection using naive Bayes classifier. In 2017 IEEE first Ukraine conference on electrical and computer engineering (UKRCON) (pp. 900-903). IEEE.
21. Sur, P., & Candès, E. J. (2019). A modern maximum-likelihood theory for high-dimensional logistic regression. Proceedings of the National Academy of Sciences, 116(29), 14516-14525.
22. Chen, S., Webb, G. I., Liu, L., & Ma, X. (2020). A novel selective naïve Bayes algorithm. Knowledge-Based Systems, 192, 105361.
23. Chaudhari, P., & Soatto, S. (2018, February). Stochastic gradient descent performs variational inference, converges to limit cycles for deep networks. In 2018 Information Theory and Applications Workshop (ITA) (pp. 1-10). IEEE.
24. Ruchansky, N., Seo, S., & Liu, Y. (2017, November). Csi: A hybrid deep model for fake news detection. In Proceedings of the 2017 ACM on Conference on Information and Knowledge Management (pp. 797-806).
25. Lee, J. Y., & Dernoncourt, F. (2016). Sequential short-text classification with recurrent and convolutional neural networks. arXiv preprint arXiv:1603.03827.
26. Wang, D., & Nyberg, E. (2015, July). A long short-term memory model for answer sentence selection in question answering. In Proceedings of the 53rd Annual Meeting of the Association for Computational Linguistics and the 7th International Joint Conference on Natural Language Processing (Volume 2: Short Papers) (pp. 707-712).
27. Sutskever, I., Vinyals, O., & Le, Q. V. (2014). Sequence to sequence learning with neural networks. Advances in neural information processing systems.
28. Sami, M., & Mobin, I. (2019, March). A comparative study on variational autoencoders and generative adversarial networks. In *2019 International Conference of Artificial Intelligence and Information Technology (ICAIIT)* (pp. 1-5). IEEE.